\documentclass[preprint,review,12pt]{elsarticle}
\usepackage{orcidlink}
\usepackage[utf8]{inputenc} 
\usepackage[T1]{fontenc}    
\usepackage{hyperref}       
\usepackage{amssymb}
\usepackage{booktabs, multirow} 
\usepackage{soul}
\usepackage{changepage,threeparttable} 
\usepackage{array, makecell}
\usepackage{amsfonts, amsmath} 
\usepackage{graphicx}
\usepackage{url}
\usepackage{xcolor}
\usepackage{tabularx,colortbl}
\usepackage[retainorgcmds]{IEEEtrantools}
\usepackage{siunitx}
\usepackage{pdflscape}

\usepackage{enumitem}
\usepackage[normalem]{ulem}
\useunder{\uline}{\ul}{}
\usepackage{setspace} 
\usepackage[utf8]{inputenc}
\usepackage[linesnumbered, ruled, vlined]{algorithm2e}
\usepackage{rotating}
\usepackage{nicefrac}
\usepackage{xspace}
\usepackage{float}
\usepackage{caption}
\usepackage{subcaption}
\usepackage{rotating, lscape, longtable, tabu, booktabs, siunitx, multirow}
\usepackage[capitalise, noabbrev]{cleveref}
\usepackage{verbatim}
\usepackage{mathtools}

\usepackage{xcolor} 
\usepackage{graphicx}
\usepackage{adjustbox}
\usepackage{pifont}
\newcommand{\xmark}{\ding{55}}%
\usepackage{lettrine}
\usepackage{makecell}

\hypersetup{
 colorlinks,
 citecolor=cyan,
 linkcolor=red,
 urlcolor=brown}

\usepackage{tkz-graph}

\usetikzlibrary{arrows,arrows.meta,shapes,decorations.pathmorphing, decorations.markings, positioning}
\let\svtikzpicture\tikzpicture
\def\tikzpicture{\noindent\svtikzpicture}

\newcommand{\uset}[1]{\ifmmode\left\{\,#1\,\right\}\else\{\,#1\,\}\fi}
\newcommand{\ulst}[1]{\ifmmode\left[\,#1\,\right]\else[\,#1\,]\fi}
\newcommand{\upar}[1]{\ifmmode\left(\,#1\,\right)\else(\,#1\,)\fi}
\newcommand{\uioc}[1]{\ifmmode\left(\,#1\,\right]\else(\,#1\,]\fi}
\newcommand{\uico}[1]{\ifmmode\left[\,#1\,\right)\else[\,#1\,)\fi}
\newcommand{\COMMENT}[1]{}

\newcommand{\ab}[1]{{\bf [\textcolor{orange}{#1}{\bf ]}}}
\newcommand{\w}{\mathbf{w}}

\def\thanks#1{\footnotemark
    \protected@xdef\@thanks{\@thanks
        \protect\footnotetext[\the\c@footnote]{#1}}%
}

\journal{Computer Speech \& Language}

\begin{document}

\begin{frontmatter}

\title{Federating Dynamic Models using Early-Exit Architectures for Automatic Speech Recognition on Heterogeneous Clients}

\author[label2]{Mohamed Nabih Ali \corref{cor1} \href{https://orcid.org/0000-0001-9132-9220}{\includegraphics[ width=0.3cm]{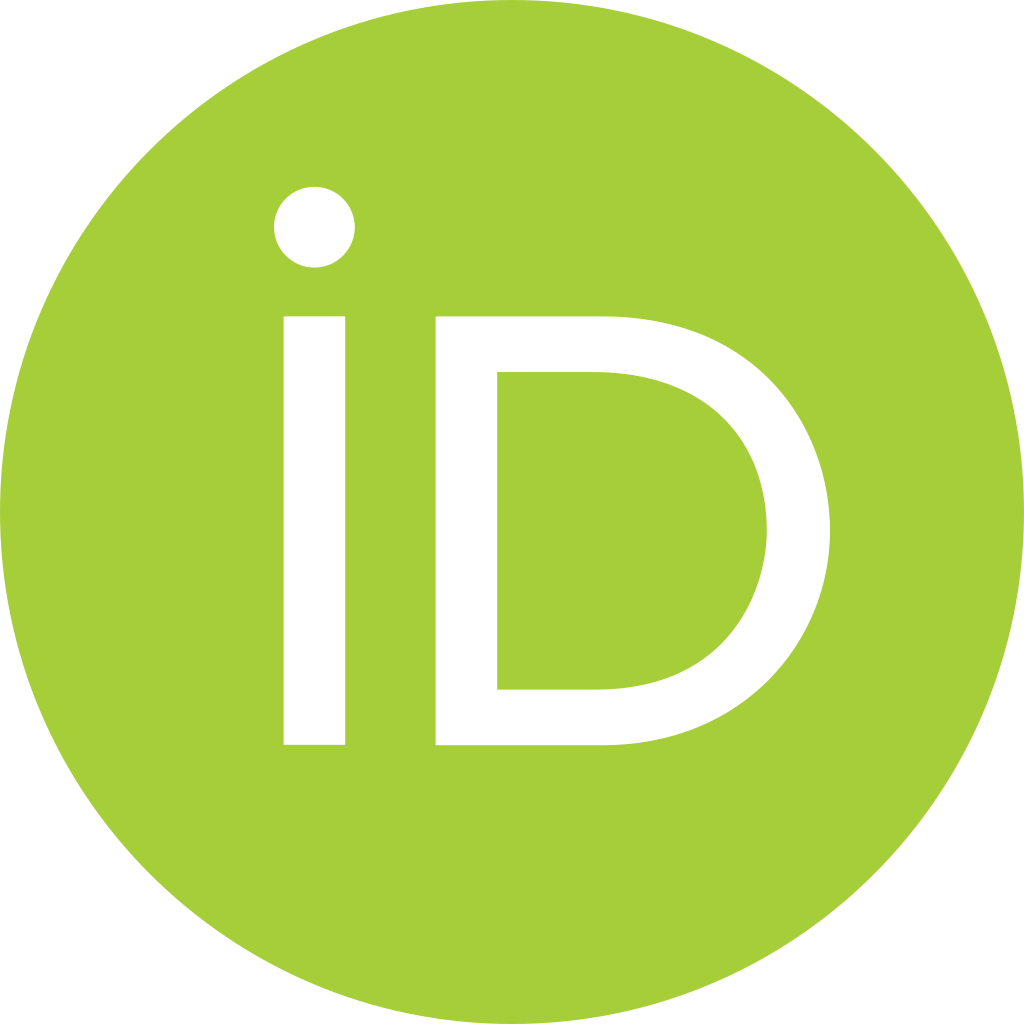}\hspace{1mm}}}
\ead{mnabih@fbk.eu}
\cortext[cor1]{Corresponding author}

\cortext[cor2]{We acknowledge the support of the PNRR project FAIR - Future AI Research (PE00000013), under the NRRP MUR program funded by the NextGenerationEU.}

\author[label2]{Alessio Brutti \href{https://orcid.org/0000-0003-4146-3071}{\includegraphics[ width=0.3cm]{orcid.png}\hspace{1mm}}}
\ead{brutti@fbk.eu}

\author[label2]{Daniele Falavigna \href{https://orcid.org/0000-0002-4844-5071}{\includegraphics[ width=0.3cm]{orcid.png}\hspace{1mm}}}
\ead{falavi@fbk.eu}

\address[label2]{Center for Augmented Intelligence, Fondazione Bruno Kessler, Trento, Italy }

\begin{abstract}
Automatic speech recognition models require large amounts of speech recordings for training. However, the collection of such data often is cumbersome and leads to privacy concerns. Federated learning has been widely used as an effective decentralized technique that collaboratively learns a shared prediction model while keeping the data local on different clients. Unfortunately, client devices often feature limited computation and communication resources leading to practical difficulties for large models. In addition, the heterogeneity that characterizes edge devices makes it sub-optimal to generate a single model that fits all of them. Differently from the recent literature, where multiple models with different architectures are used, in this work, we propose using dynamical architectures which, employing early-exit solutions, can adapt their processing (i.e. traversed layers) depending on the input and on the operation conditions. This solution falls in the realm of partial training methods and brings two benefits: a single model is used on a variety of devices; federating the models after local training is straightforward. Experiments on public datasets show that our proposed approach is effective and can be combined with basic federated learning strategies. 
\end{abstract}


\end{frontmatter}


\section{Introduction} 

\begin{figure}[htb!]
  \centering
  \includegraphics[width=\textwidth]{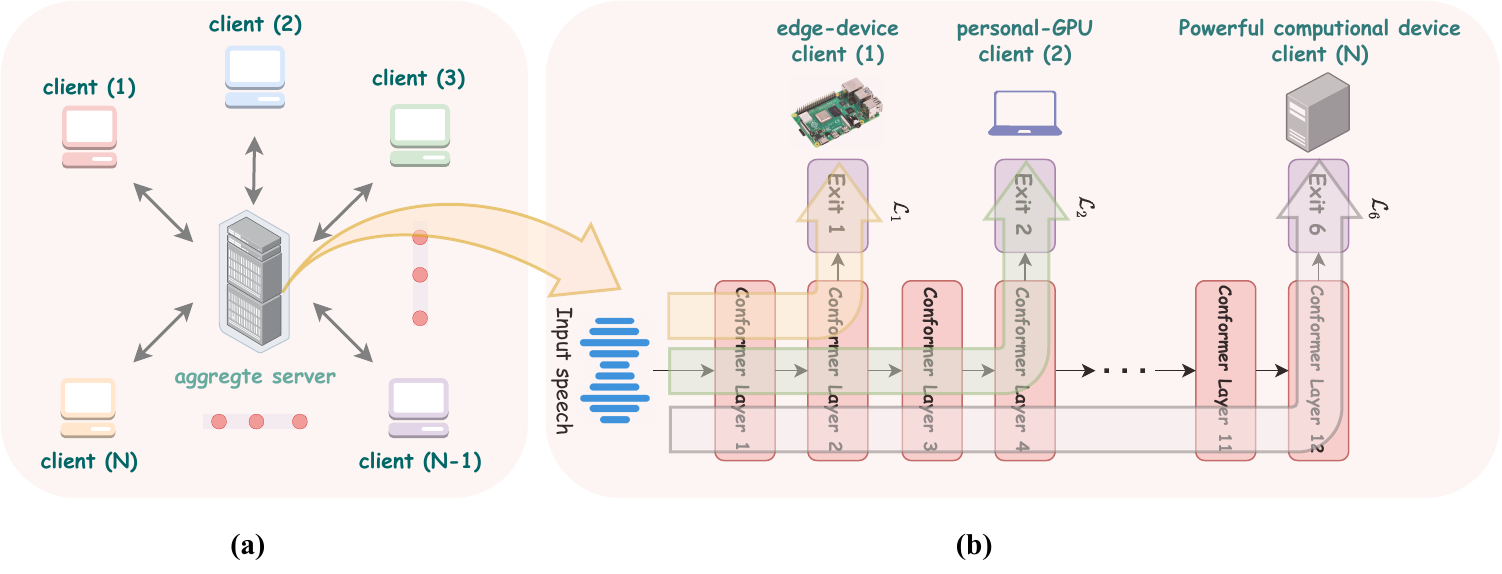}
  \caption{(a) Federated learning framework with \textbf{N} connected clients. (b) An early-exit model architecture deployed on different devices.} 
  \label{fig:FL}
\end{figure}

Nowadays, automatic speech recognition (ASR) systems based on deep learning provide excellent performance for many languages and allow to develop either products or services in many application domains~\cite{mehrish2023review, kumar2018survey}. However, the amount of memory and computation resources for handling ASR models is usually high and training is carried out on high-performing servers using centralized datasets ~\cite{rao2023federated}. Nowadays, it is common practice to fine-tune large pre-trained ASR models using data collected on the field. 
The latter requirement, in particular, poses issues related to data ownership, latency, and cost. These aspects gained a lot of attention with the widespread employment of portable devices: computers,  smartphones, wearable devices, etc., most of which featuring limited computational resources \cite{zhu2022device,Paissan2022phinets}. 
Consequently, training over a distributed environment, a strategy known as federated learning (FL), received growing attention from the scientific community giving rise to a flourish of investigations and research~\cite{mcmahan2017communication, matsubara2022split, gao2022end, li2020review}. The main aim of FL is to effectively utilize the abundant local private data by training the model locally without the need for a centralized huge dataset. As depicted in Fig. \ref{fig:FL}(a), this is achieved by combining pieces of information from the connected edge devices, while keeping the data decentralized. 

In practical application scenarios, besides the typical FL issues related to the variability of data distributions across clients, the heterogeneity and volatility of the computational resources available on the clients introduce a further challenge. In most FL methodologies, a foundational assumption entails treating all clients as "homogeneous" in terms of their computational assets, presuming compatibility with a common shared model architecture. However, this assumption encounters limitations in practical scenarios where devices exhibit substantial disparities in memory capacity, computational capabilities, and power consumption. In particular, the available {\it resources are volatile}, depending on what other tasks the device may be handling. Moreover, real-world application domains frequently involve the presence of low-end devices constrained by limited resources. Consequently, the overarching FL framework necessitates managing diverse "heterogeneous" architectures, guided by personalized tasks and local resource constraints~\cite{tan2022towards}. In the following, we define Homogeneous FL (OFL) the condition all clients share the same model architecture. Conversely, we denote with Heterogeneous FL (EFL) the case in which different architectures are deployed on the clients. In general, OFL is simpler than EFL, since centralized updates affect all the parameters of the same shared model, while in EFL different architectures have to be managed. The recent state of the art tackles the EFL scenario by: {\em a)} maintaining multiple central models (one for each different architecture) and/or {\em b)} implementing articulated agglomeration strategies, e.g. by sharing some layers across the various centralized models~\cite{park2023fedhm, cho2022flame, cho2022heterogeneous}. Although effective in terms of performance, these approaches require maintaining and federating multiple different models, which demand inevitably more resources (in terms of storage and computation). As an alternative, methods based on partial training are recently emerging~\cite{, diao2021heterofl,Jiang2023PruneFL, alam2022fedrolex,guliani2022dropout}, which employ sub-nets or pruning to obtain partial networks of different dimensions that fit the client resources. 
Following this direction, 
 we propose to use early-exit (EE) architectures, which introduce intermediate exit branches~\cite{teerapittayanon2016branchynet, phuong2019distillation} (see Fig.~\ref{fig:FL}(b)) so that an output can be made available after a subset of the layers has processed the input. This results in multiple scaled versions of the same architectures, that can fit different (time-varying) client requirements. If properly trained, EE models can provide extremely good ASR performance in the lower exits compared against the full model~\cite{wright2023training}. In section~\ref{sec:FLwithEE} we show that EE architectures allow agglomerating coherently the model updates at the server side.
Moreover, being a partial training strategy thanks to the modularity of EE architectures, our approach naturally blends with differential privacy \cite{dwork2014algorithmic} and secure aggregation \cite{bonawitz2016practical} solutions (as discussed in~\cite{alam2022fedrolex}), two techniques often applied to improve the protection of private data in FL settings.

This paper extends our previous contribution in \cite{nawar2023fed} with better mathematical grounding and contextualization in the state of the art of our proposed approach, together with further experimental insights and ablations. Overall, we aim to contribute to the scientific literature in the following three directions:
\begin{itemize}
    \item We provide a mathematical analysis of EE in FL, proving that FL training a set of heterogeneous EE models (i.e. with different numbers of processing layers) is equivalent to training a single homogeneous EE model, if appropriately combining the losses of the various exits.
    \item We show that EE architectures allow federating heterogeneous models in a rather straightforward way. This surpasses current approaches based on the use of multiple models centrally aligned with edge-specific solutions \cite{cho2022heterogeneous}, also allowing the combined use of differential privacy and secure aggregation.
    \item We shed lights on some implementation aspects of federated ASR, showing the efficacy of the FedAdam agglomeration strategy \cite{reddi2020adaptive} and of freezing part of the pre-trained model (pre-trained in an EE fashion).
\end{itemize}

\section{Related Work}
In this section we provide a revision of the state of the art considering first of all FL applied to ASR and the related common practices. We then revise techniques for FL in presence of heterogeneous devices and architectures and finally we comment on the implications of EFL in current methods for differential privacy.
\subsection{Federated learning for ASR}
\label{sec:FLASR}
Besides the typical issues related to non-i.i.d and unbalancing data distributions, FL for ASR is further complicated by the high computational requirements of most ASR architectures (e.g Transformers \cite{zeyer2019comparison}, Transducers \cite{moriya2023improving, zeineldeen2022conformer} and RNNs \cite{oruh2022long}), which  may be not available on clients, and the need for large datasets~\cite{wang-etal-2021-voxpopuli}, usually featured only in the central servers.
A further complication is the lack of ground truth across the clients which demands for unsupervised or self-supervised approaches.
Recently several papers have been published on FL for ASR. In the following we report the most relevant in our opinion. A deep exploration of several optimization approaches and training strategies 
is available in~\cite{azam2023importance, yu2021federated}. We forward the reader to those publications for a more comprehensive survey and some practical suggestions.

A pioneer work on FL in ASR is \cite{dimitriadis2020federated} which proposes two different dynamic aggregation methods for the gradients. Client-specific weighting strategies are frequently employed in the agglomeration stage, as in \cite{gao2022federated} which explores both loss-based and word error rate-based weighting, demonstrating the superior performance of the latter. \cite{gao2022federated}
also underscores the necessity of pre-training the model centrally (before FL) to achieve convergence. Additionally, the paper introduces a central training stage, subsequent to aggregation, to regulate model divergence between consecutive rounds. Similar observations are reported in \cite{gao2022end}. Note, however, that \cite{nguyen2023federated}, using the TED-LIUM-3 dataset~\cite{hernandez2018ted}, partially diverges from these observations, asserting that only the use of a large pre-trained model (Wav2Vec2.0 \cite{baevski2020wav2vec}) is effective for cross-domain ASR tasks. 

Differently from these findings, we show that our proposed EE model achieves performance improvements in FL on TED-LIUM-3 and VoxPopuli, even when starting from a global model trained solely on Librispeech~\cite{panayotov2015librispeech}. Furthermore, we observe that the use of central training on a held-out training set does not bring particular benefits. 

Finally, motivated by the lack of labels typical of FL settings, \cite{jia2022federated} and \cite{dimitriadis2020federated} investigate self-supervised and unsupervised FL approaches. Since our work proposes the use of a specific neural architecture for EFL, we do not address here solutions to handle the lack of supervision. Therefore, 
general solutions for self- or unsupervised training may be employed.

\subsection{Federated Learning with heterogeneous devices}
\label{sec:fedhetero}
\begin{figure}[ht]
 \centering
 \includegraphics[width=14cm]{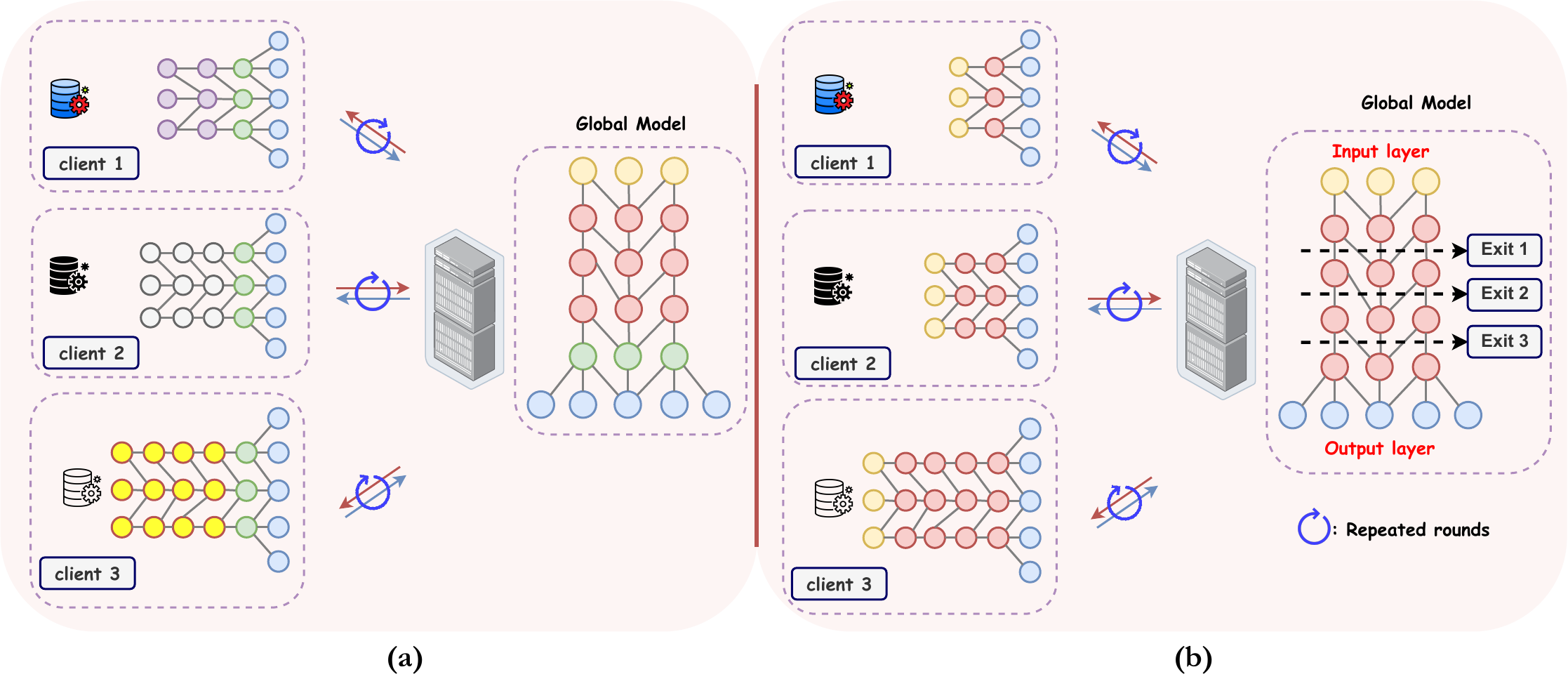}
 \caption{Illustration of server-client aggregation strategy with heterogeneous models. (a) In the common SOTA approaches clients' models have different architectures, sharing a common part (azure nodes) with the global model \cite{cho2022heterogeneous}. (b) Our proposed approach where all clients share the same architecture (with different layers according to computational resources), while their central agglomeration results into an EE model.}
 \label{fig:agg}
\end{figure}

FL with heterogeneous devices and architectures is receiving growing interest in literature recently, also from a theoretical perspective~\cite{zhou2023every}. Recent works have addressed this challenge primarily by considering multiple neural architectures with distinct characteristics, yet sharing certain common components, as illustrated in Fig.~\ref{fig:agg}(a). 


One example is presented in~\cite{mills2021multi}, where mixed architectures are adopted for local models, sharing a subset of parameters with the central model. To preserve global model parameters while assimilating local information, knowledge distillation is employed at the client side~\cite{ni2022federated, lin2020ensemble}. Alternatively, a contrastive loss mechanism can be used to minimize the dissimilarity between central and local heterogeneous models~\cite{li2021model}. Finally, the Federated Ensemble Knowledge Transfer (Fed-ET)~\cite{cho2022heterogeneous} employs consensus distillation derived from clients, enabling the regulation of divergence from a large central model.

Training and managing multiple models with varying architectures is complex and computationally inefficient. Recently, approaches based on the concept of partial training (PT) are emerging, where the central models are somehow decomposed in different subnetworks to be deployed on the clients and that can be easily agglomerated. Federated Dropout \cite{guliani2022dropout} and FjORD \cite{horvath2021fjord} handles FL for heterogeneous edge devices selecting lower footprint sub-models via drop-out. 
HeteroFL~\cite{diao2021heterofl} proposes an analogous approach but focusing on the width of the hidden channels. 
PruneFL~\cite{Jiang2023PruneFL} propose dynamically adjustable pruning strategies of the models to improve the efficiency and reduce the communication costs, keeping the same pruned model for all clients. Finally, FedRolex~\cite{alam2022fedrolex} applies an analogous approach but also considering different pruned models in the edge devices.


Note that, with respect to \cite{guliani2022dropout, horvath2021fjord, diao2021heterofl, Jiang2023PruneFL}, our approach, thanks to exit-specific decoders of EE architectures, optimizes the performance at the level of each early exit with a layer-specific loss, providing higher robustness to heavy model reduction compared to layer-drop and pruning.

To conclude, it is worth mentioning that most of the approaches for EFL reported here have demonstrated effectiveness for image or text processing, leaving several speech-processing issues unresolved.

\subsection{Differential Privacy}
Several works have recently shown that in FL settings information about the data used for local training can be inferred from the gradients or the updated weights. 
For ASR only preliminary analyses are available: for example \cite{nguyen2023federated, tomashenko2022privacy} demonstrates that the speaker identity can be retrieved while there are no proofs that the acoustic features or the acoustic content can be obtained. On the contrary, these privacy attacks have been fully demonstrated for text and image processing~\cite{boenisch2023curious}.

To actually preserve privacy in distributed model environments Differential Privacy (DP) \cite{dwork2014algorithmic} and/or Secure Aggregation (SA) protocols \cite{bonawitz2016practical} have been introduced. Towards DP, large models \cite{shen2021towards} as well as large numbers of active clients in each round should be employed.  This is somehow in contrast with the realistic requirements of FL for ASR, where many devices cannot run large models and where some rounds can involve only few clients. Some preliminary works have applied foundations of DP to ASR in distributed setting, providing performance figures on standard benchmarks. 
For example, \cite{shoemate2022sotto} applies DP and FL for senone classification, 
while \cite{pelikan2023federated} sets-up useful reference baselines propossing per-layer clipping of gradients to compensate for the adversarial noise introduced by DP. 

To conclude, although DP and SA are nowadays indispensable tools to be coupled with FL, in this work we do not explicitly consider their implementations because our PT-based proposed approach is compliant, as reported in ~\cite{alam2022fedrolex}, with the methods reported above.

\section{Proposed Approach}
\label{sec:proposed}
\subsection{Federated Learning Framework}
\label{subsec:FLframework}
The goal of FL is to optimize a global model, on a server, through a sequence of rounds each agglomerating the parameters sent by the connected clients.
Let us assume that a set of clients $\mathcal{C}$ are available at round $\tau$, and each client $c \in \{1,...,|\mathcal{C}|\}$ observes its own training set $\mathcal{B}_c$. At the beginning of each round, the clients receive the most recently trained model, $\w(\tau)$  from the central server: 
\begin{equation}
    \w^c(\tau,t)=\w(\tau), \ \ \ (t=1)
\end{equation}
 The $k$-th parameter ($k\in \{1, \dots,K\}$) of each model is updated for a certain number of epochs $E \ (1<t\leq E)$ applying SGD to the local dataset $\mathcal{B}_c$ and providing "local gradients" $\nabla \w_k^c(\tau)=\w_k^c(\tau,E)-\w_k^c(\tau,1)$ to the server (we use the time indexes $t$ and $\tau$ for local iterations/epochs and server agglomeration rounds, respectively). Note that in the case of homogeneous devices, all clients accommodate the same network.
A common approach to update the global model parameters is by federated averaging the locally computed gradients $\nabla \w_k^c(\tau)$ (FedAvg ~\cite{mcmahan2017communication}),
 possibly applying clients specific weights (as in in~\cite{gao2022end}), as follows:
\begin{equation}
\label{eq:fedavg}
    \nabla \w_k(\tau)=\sum_{c\in\mathcal{C}}\gamma_c(\tau) \nabla \w_k^c(\tau)
\end{equation}
where the weighting coefficients $\gamma_c(\tau)$ estimates the client confidence, considering the size of the local training data or client-specific validation metrics.

In the experiments reported below we  apply FedAdam~\cite{reddi2020adaptive}: a global adaptive optimization step, applied to the "pseudo-gradient" $\nabla \w_k(\tau)$, which has demonstrated to increase the convergence speed of FL on many tasks. 



\subsection{Federated Learning with Early-Exit models}
\label{sec:FLwithEE}
In the presence of devices with different processing capabilities, using a single model $\w$ is not feasible. Most of current approaches employ $U$ different models $\w^u$, $u\in\{1,\dots,U\}$ with different memory and computation requirements which are all maintained on the centralized server. 
As mentioned in Section~\ref{sec:fedhetero}, an interesting solution is offered by EE architectures which provide $M$ exit layers producing hypothesis: $\hat{\mathbf{y}}^1,\ldots,\hat{\mathbf{y}}^M$. The overall model is trained by optimizing the joint objective~\cite{teerapittayanon2016branchynet, wright2023training}: 
\begin{equation}
\mathcal{L}_{EE}(\hat{\mathbf{y}}^1,\ldots,\hat{\mathbf{y}}^M,\mathbf{y}) = 
\sum_{m=1}^M\mathcal{L}_m(\mathcal{B}),
\label{eq:EEloss}
\end{equation}
where $\mathcal{L}_m$ is the loss at the $m$-th exit layer, $\mathbf{y}$ is the ground-truth and $\mathcal{B}$ is the overall training set. 
In a standard FL framework where the same architecture (i.e. the full EE network) is deployed in all clients, Eq.~\ref{eq:fedavg} becomes:

\begin{equation}
 \nabla \w_k(\tau)=\frac{1}{|\mathcal{C}|}\sum_{c\in \mathcal{C}} \sum_{m=1}^{M}\nabla_k^{c,m}(\tau)=\frac{1}{|\mathcal{C}|} \sum_{m=1}^{M}\sum_{c\in\mathcal{C}}\nabla_k^{c,m}(\tau)
\label{eq:EE-FEDL}
\end{equation}
where $\nabla_k^{c,m}(\tau)$ represents the local gradient of client $c$ at exit $m$. 
For the sake of simplicity, we assume that $\gamma_c(\tau)= 1/{|\mathcal{C}|}$. Basically, each sub-network, corresponding to a given exit $m$, is centrally updated averaging all the corresponding sub-nets on the clients, leading to a \emph{per-exit} agglomeration.

\subsection{Federated Learning with Early-Exits on heterogeneous devices}
Let us assume that the EE model on the server, $\w$, is split into $M$ sub-nets, $\w(l_1),\ldots, \w(l_M)$, each of them corresponding to an exit layer, where $l_m$ represents highest exit layers of the $m$-th sub-net and $l_1\le l_2\le\ldots \le l_M$ (these are our $U$ $\w^u$ models in~\cite{cho2022heterogeneous} for example).
In the heterogeneous device scenario, each client $c$ fits a sub-net $l_m$ with probability $\pi(m)$. We denote as $\mathcal{C}_m$ the set of clients with sub-net $l_m$, such that $\sum_{m=1}^{M} |\mathcal{C}_m|=|\mathcal{C}|$ (note that in this way $\pi(m)$ can be estimated as: $\pi(m)={|\mathcal{C}_m|}/{|\mathcal{C}|}$). 

Following the notation above, we also denote with $\w_k(\tau,l_m)$ the $k$-th parameter of the EE sub-net with $l_m$ exits at FL round $\tau$. Each client $c$, exhibiting $l_m$ exits,  locally updates the model applying SGD to the compound loss (see Eq. \ref{eq:EEloss}) providing at round $\tau$ the following local gradients:
\begin{equation}
    \nabla\w_k^c(\tau,l_m) = \sum_{i=1}^{l_m}\nabla_k^{c,i}(\tau).
    \label{eq:localgradient}
\end{equation}
%

Therefore, following Eq.~\ref{eq:EE-FEDL}, the FedAvg agglomerated gradients from all clients, fitting models with $l_m$ exits,  are:
textcolor{red}{
\begin{equation}
\nabla \w_k(\tau, l_m) = \frac{1}{|\mathcal{C}_m|}\sum_{c\in\mathcal{C}_m} \sum_{i=1}^{l_m}\nabla_k^{c,i}{(\tau)}  = \frac{1}{|\mathcal{C}_m|}\sum_{i=1}^{l_m} \sum_{c\in\mathcal{C}_m}\nabla_k^{c,i}(\tau)
\label{eq:clientindipendentgradient}    
\end{equation}
Note that if all clients fit the same full model $\w(\tau,l_M)$, $\mathcal{C}_m=\mathcal{C}$ and the equation above becomes Eq.~\ref{eq:EE-FEDL}, which is the standard FedAvg formulation for EE architectures. Eq.~\ref{eq:clientindipendentgradient} updates in a FL fashion the weights of the $m$-th sub-net, updating all the exits from 1 to $l_m$. If we now aggregate all the sub-nets, we obtain the global gradients as sum of all the model-dependent gradients: 
\begin{equation}
    \nabla \w_k(\tau) =  \sum_{m=1}^{M} \nabla \w_k(\tau, l_m),
    \label{eq:ggrad}
\end{equation}
where the gradients for parameters $k$ not in $\w(\tau,l_m)$ is obviously 0. Replacing Eq.~\ref{eq:clientindipendentgradient} into Eq.~\ref{eq:ggrad} we obtain:
\begin{align}
\nonumber\nabla \w_k(\tau) &=   \sum_{m=1}^{M}\frac{1}{|\mathcal{C}_m|}\sum_{i=1}^{l_m} \sum_{c\in\mathcal{C}_m}\nabla_k^{c,i}(\tau)
= \sum_{m=1}^{M}\frac{1}{|\mathcal{C}_m|}\sum_{c\in \cup_{i=1}^m \mathcal{C}_i} \nabla_k^{c,m}(\tau)\\
&=\frac{1}{|\mathcal{C}|}\sum_{m=1}^{M}\frac{1}{\pi(m)}\sum_{c\in \cup_{i=1}^m \mathcal{C}_i} \nabla_k^{c,m}(\tau)
    \label{eq:EEaggregatedGradient}
\end{align}

Note that Eq.~\ref{eq:EEaggregatedGradient} resembles Eq.~\ref{eq:EE-FEDL}, which is the FedAvg formulation for EE architectures, except that a subset of clients is used for some of the layers, in particular the upper ones. These results in a weighted combination of the $M$ EE losses based on the number of clients fitting the different sub-nets $\w(l_m)$. Therefore, the equation above can be approximated as a weighted version of Eq.~\ref{eq:EE-FEDL} (see also the example below for a clarification):

\begin{equation}
   \nabla \w_k(\tau) = \frac{1}{|\mathcal{C}|}\sum_{m=1}^{M}\frac{1}{\pi(m)}\sum_{c\in \cup_{i=1}^m \mathcal{C}_i} \nabla_k^{c,m}(\tau) \approx \frac{1}{|\mathcal{C}|} \sum_{m=1}^{M} \xi_m \sum_{c\in \mathcal{C}} \nabla_k^{c,m}(\tau),
\end{equation}

where the weighting parameter $\xi_m$ is a function of the number of clients in each set $\mathcal{C}_m$. Keeping in mind that lower layers are present in all sub-nets fitting higher layers (e.g. layer 1 is always present), we have

\begin{equation}
    \xi_m= \left\{
    \begin{array}{ll}
         \frac{M-m}{\pi(m)} & |\mathcal{C}_m| > 0 \\ \\
          0  & |\mathcal{C}_m|=0\\
    \end{array} 
    \right.               
\label{eq:gamma}
\end{equation}

As a consequence, although heterogeneous architectures are non-uniformly distributed along the various FL rounds (e.g. some of them can lack in round $\tau$, while others can be present many times), we can assume for the gradients an expression formed by the weighted sum of single-exit gradients
Overall, using non-homogeneous EE architectures results in {\it layer-dependent learning rates} that slow down the convergence of the less represented layers but without compromising the convergence, as experimentally shown in section~\ref{sub:expenonuniform}:
\begin{equation}
    \eta_m=\xi_m\eta \sim \frac{M-m}{\pi(m)}\eta
    \label{eq:m-dependent-lr}
\end{equation}

\subsection{Example of heterogeneous devices}

For the sake of clarity, we now report a simple practical example. 
In a OFL scenario, one central EE model with $M$ exits is trained over $|\mathcal{C}|$ clients (assume without losing generality a fixed number of clients per round). The local gradients from every exit equally contribute to the agglomerated gradient of Eq.~\ref{eq:ggrad}. Instead, in a EFL scenario, the central EE model agglomerates gradients from models with different number of layers: clients fitting more layers contribute to FedAvg with both the gradients of the upmost exits and the lowest ones. As a result, the gradients of the lowest layers are weighted more   in the central EE model.

For example, consider the case, similar to that  depicted in Figure~\ref{fig:agg}b),
where 3 EE clients  $\w^a$, $\w^b$, and $\w^c$ with 1, 2 and 3 exits, respectively, are available.
At round $\tau$ the gradients of the centralized model $\w_k(\tau)$, according to Eq.~\ref{eq:clientindipendentgradient} and~\ref{eq:ggrad} are: 
\newline
$\nabla\w_k(\tau) = \nabla\w^{a,1}_k(\tau) + \nabla\w^{b,1}_k(\tau)  + \nabla\w^{b,2}_k(\tau) + \nabla\w^{c,1}_k(\tau) + \nabla\w^{c,2}_k(\tau) + \nabla\w^{c,3}_k(\tau)$
Note that exit 1 is present three times, exit 2 twice and exit 3 only once. This is equivalent, in Eq.~\ref{eq:ggrad}, weighting 3 the client $a$, 2 the client $b$ and 1 the client $c$, i.e.$\nabla\w_k(\tau)=3\times\nabla\w_k(\tau,1) + 2\times\nabla\w_k(\tau,2) + \nabla\w_k(\tau,3)$.  
Therefore,  the lower sub-nets result enforced after their agglomeration, as if they occur more times in the overall distributed architecture.

\COMMENT{
by agglomerating the gradients coming from all the three clients, i.e. using the notation introduced above: $\nabla \w_k(\tau,2)=\sum_{i=1}^{i=2}(\nabla \w^{1,i}_k(\tau) + \nabla \w^{2,i}_k(\tau) + \nabla \w^{3,i}_k(\tau))$. On the contrary the gradients coming from $\w^1$ do not contribute to form the agglomerated gradients of exits 3 and 5 in the centralized EE model,   and  gradients of both $\w^1$ and $\w^2$ do not contribute to the agglomerated gradient of exit 5 (i.e. $\nabla \w_k(\tau,3)=\sum_{i=1}^{i=3}(\nabla \w^{2,i}_k(\tau) + \nabla \w^{3,i}_k(\tau))$, $\nabla\w_k(\tau,5)=\sum_{i=1}^{i=5}\nabla \w^{3,i}_k(\tau)$. Therefore,  the lower sub-nets result enforced after their agglomeration, as if they occur more times in the overall distributed architecture.}

\section{Experimental Setup}
We evaluate our proposed FL framework on TED-LIUM-3~\cite{hernandez2018ted} and on the English part of VoxPopuli~\cite{wang-etal-2021-voxpopuli}. TED-LIUM-3 includes English TED talks, for a total of 452 hours of speech from 1938 speakers. VoxPopuli is a large-scale multilingual corpus with 1.8K hours of speech. In our experiments, we use the English subset from 1313 speakers.

Following the best practice in literature and in an attempt to make the scenario as realistic as possible, the training set of TED-LIUM-3 and Voxpopuli is split in such a way that each client sees data of a single speaker (this way mimicking personal devices). As observed in previous studies~\cite{nguyen2023federated}, training an ASR model from scratch in a federated fashion is not feasible. Therefore we {\bf pre-train} our model in a centralized way using the whole 960 hours of the Librispeech training set~\cite{panayotov2015librispeech}. Note that, differently from the current literature, this choice implies that we use FL for {\bf domain adaptation} rather than for fine-tuning on the same domain. 
Performance is measured in terms of WER on each of the $M$ exits of the resulting agglomerated model. 

For the {\bf ASR model}\footnote{The code is available at: \url{https://github.com/augustgw/early-exit-transformer}}, we use the EE architecture shown in Fig.~\ref{fig:FL}(b)~\cite{wright2023training}. The network takes as input 80 Mel Frequency Cepstral Coefficients (MFCCs). This sequence is passed through two depth-wise convolutional layers, each downsampling the input data by factor $2$, followed by a stack of $N=12$ conformer layers with $M=6$ intermediate decoders (one every 2 conformer layers, $M=\frac{N}{2}$). Each exit is an independent head consisting of a linear layer. The model uses a byte pair encoding (BPE) based tokenizer \cite{sennrich2015neural} with 256 tokens and it is trained using the CTC loss \cite{graves2014towards,graves2013speech}. 
Table \ref{tab:hyperparams} summarizes the main hyperparameters of the model and of the training procedure.

\begin{table}
\centering
\scriptsize
\caption{Hyperparameters for the EE architecture shown in Fig.\ref{fig:FL}(b).}\label{tab:hyperparams}
\begin{tabular*}{\textwidth}{c @{\extracolsep{\fill}} cccccccccc}
\toprule
\makecell{Para. \\ (M)} & \makecell{Enc. \\ layers} & \makecell{Att. \\ dim.} & Heads \# & \makecell{FF \\ dim} & \makecell{Dec. \\ type} &Input & Loss & \makecell{Output \\ units} & \makecell{ LM \\ scoring} \\ 
\midrule
31 & 12-layers & 256 & 8 & 2048 & Linear & 80-MFCC & CTC & BPE & \xmark \\
\toprule
\end{tabular*}
\end{table}

We implement the {\bf FL framework}\footnote{The code is available at: \url{https://github.com/mnabihali/ASR-FL}} using Flower \cite{beutel2020flower}. As mentioned above, we deploy a client for each speaker: 1938 for TED-LIUM-3 and 580 for VoxPopuli. In each round,  $10\%$ of available clients are randomly instantiated for local training, which applies SGD for $E=5$ epochs with learning rate 0.01. These values were empirically chosen to achieve learning efficacy on local data while avoiding overfitting. Models are centrally agglomerated using either \textbf{FedAvg} or \textbf{FedAdam}. In order to speed up the convergence we also consider \textbf{freezing the convolutional front-end}, as suggested in~\cite{yu2021federated}.

Finally, we consider two FL scenarios: OFL) the traditional one, where {\bf homogeneous} models are deployed on the clients (i.e. the full EE architecture), that is our FL upper-bound; EFL) the case where devices are {\bf heterogeneous} and the number of available exits randomly varies across clients and rounds. For the latter case, which is the scenario of interest in our work, we simulate 3 conditions: 1) {\it uniform} distributions of the sub-nets on the clients (i.e. each sub-net has probability $\frac{1}{M}$); 2) {\it non uniform} distributions where smaller sub-nets are more likely; 3) {\it extreme non uniform} distribution where highest exits are rarely available. 

\subsection{Experimental Results}
\begin{table}\centering
\caption{Three leftmost columns: WER of the model pre-trained on Librispeech and evaluated on Librispeech, TED-LIUM-3 and VoxPopuli. The 4th and 5th columns report the FL performance using FedAdam and the frozen convolutional front-end on TED-LIUM-3 and VoxPopuli. The last two columns report the upper-bound performance when training on TED-LIUM-3 and Voxpopuli in a centralized way.}
\label{tab:centralizedTranining}
\scriptsize
\begin{tabular}{c ccccc cc}\toprule
 &\multicolumn{3}{c}{\textbf{Librispeech-960$^{*}$}} & \multicolumn{2}{c}{\textbf{FedAdam-Hetero-Freeze}} & \textbf{TED-LIUM-3$^{\dagger}$} & \textbf{Voxpopuli$^{\dagger}$} \\\cmidrule{2-8}
\textbf{Exit} & \makecell{Libri \\ Test-Clean} & \makecell{TED \\Test} & \makecell{Vox\\Test} &\makecell{TED-Test\\1650 rounds} & \makecell{Vox-Test\\ 4000 rounds} & \makecell{TED\\Test}& \makecell{Vox\\Test}\\\cmidrule(l{2pt}r{2pt}){2-4} \cmidrule(l{2pt}r{2pt}){5-6}\cmidrule(l{2pt}r{2pt}){7-8}
\textbf{1} & 24.10 & 50.94 & 62.91 & 45.20 & 43.36 & 43.8 & 36.7 \\\cmidrule{1-8}
\textbf{2} & 11.78 & 34.50 & 40.10 & 29.92 & 26.73 & 23.4 & 21.1 \\\cmidrule{1-8}
\textbf{3} & 6.91  & 29.58 & 33.17 & 24.16 & 20.95 & 18.0 & 17.3 \\\cmidrule{1-8}
\textbf{4} & 6.28  & 28.61 & 31.57 & 23.22 & 19.64 & 16.1 & 15.4 \\\cmidrule{1-8}
\textbf{5} & 7.30  & 28.79 & 30.89 & 23.13 & 18.98 & 14.9 & 14.7\\\cmidrule{1-8}
\textbf{6} & 5.34  & 27.35 & 30.26 & 21.70 & 18.06 & 14.6 & 14.3 \\
\bottomrule \\
\multicolumn{8}{l}{\footnotesize $^{*}$ This model was used as an initial starting model for all federated learning experiments.} \\
\multicolumn{8}{l}{\footnotesize $\dagger$ Upper-bound WER with centralized training.} \\
\end{tabular}
\end{table}

The results of our experimental analysis are reported in Table~\ref{tab:centralizedTranining} in terms of WER for each exit, comparing our proposed FL approach for heterogeneous devices with central training. The three leftmost columns report the performance of the model pre-trained on Librispeech. The first column shows the performance on the Librispeech test-clean set, confirming that our architecture is a reasonable baseline in comparison with the state of the art (note that in order to save computation time we have adopted a quite small model in the experiments and we didn't apply data augmentation in training). The second and third columns report the WER achieved by the Librispeech pre-trained model on TED-LIUM-3 and VoxPopuli. This is our starting point, on which we apply our FL approach. The last two columns reports our upper-bound, given by the model trained from scratch in a centralized way. Finally, the results of our FL approach using heterogeneous models with FedAdam and freezing the convolutional layers are reported in the fourth and fifth columns. Note that, using the EE architecture, even if not all the subnets are always available at all clients, we can improve the model performance towards the new domain for all exits, approaching the performance of the model centrally trained. Due to the heavy computational cost of simulating a federated infrastructure, we could not achieve full convergence of the models, however the trends are clear.
In the following sections, we provide more details on some specific aspects of our framework.

\subsection{Heterogeneous vs Homogeneous devices}

\begin{figure}[ht!]
  \centering
       \includegraphics[width=\textwidth]{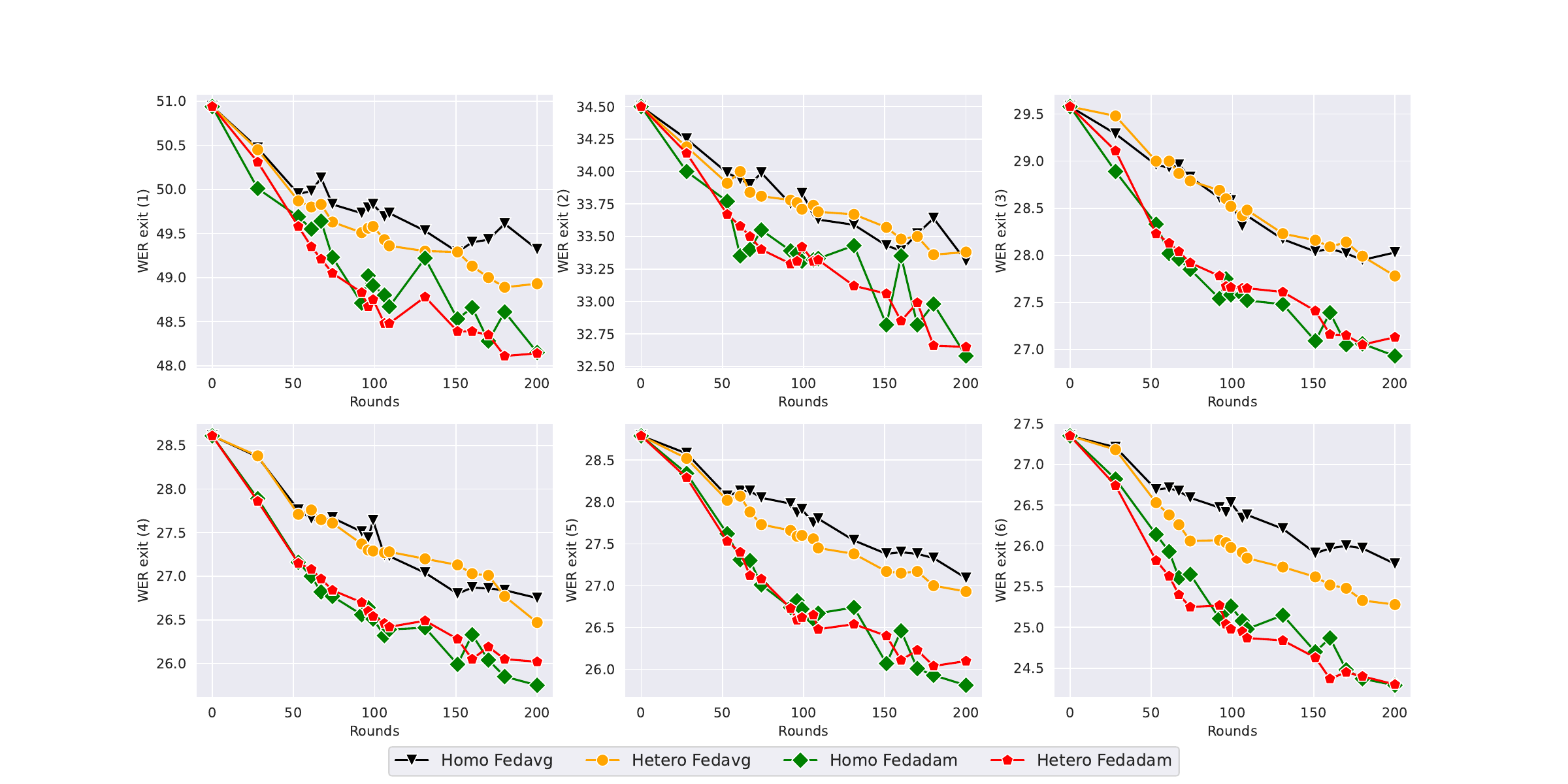}
    \caption{WER achieved on TED-LIUM-3 with homogeneous and heterogeneous devices, using FedAvg and FedAdam during the first 200 FL rounds. Each figure refers to one exit. Heterogeneous models are uniformly distributed.}
  \label{fig:strat}
\end{figure}

The fundamental claim of our work, mathematically demonstrated in Section~\ref{sec:FLwithEE}, is that using EE architectures we can train the full model even if only part of that is available in some of the clients due to device heterogeneity. In Fig.~\ref{fig:strat} we provide more details on the proposed approach, comparing WER trends of both OFL and EFL along the FL rounds for all the 6 exits (each sub-figure refers to a single exit, top left is the lowest exit, bottom right the last one), aggregating either with FedAdam or FedAvg. Note that in this case the heterogeneous models are uniformly distributed on the clients. The first interesting result is that agglomerating the models with standard FL strategies (i.e. with both FedAvg and FedAdam) is viable, for all exits, even in EFL conditions. Note that the performance for both homogeneous and heterogeneous models are very similar for all exits and for both agglomeration methods. This confirms the efficacy of EE architectures in this scenario as well as our claim. Furthermore, similarly to~\cite{ju2023accelerating}, the figure clearly exhibits the superior performance of FedAdam (green and red lines) with respect to FedAvg (black and yellow), both in terms of WERs and convergence speed. 
\begin{figure}[ht!]
  \centering
      \includegraphics[width=\textwidth]{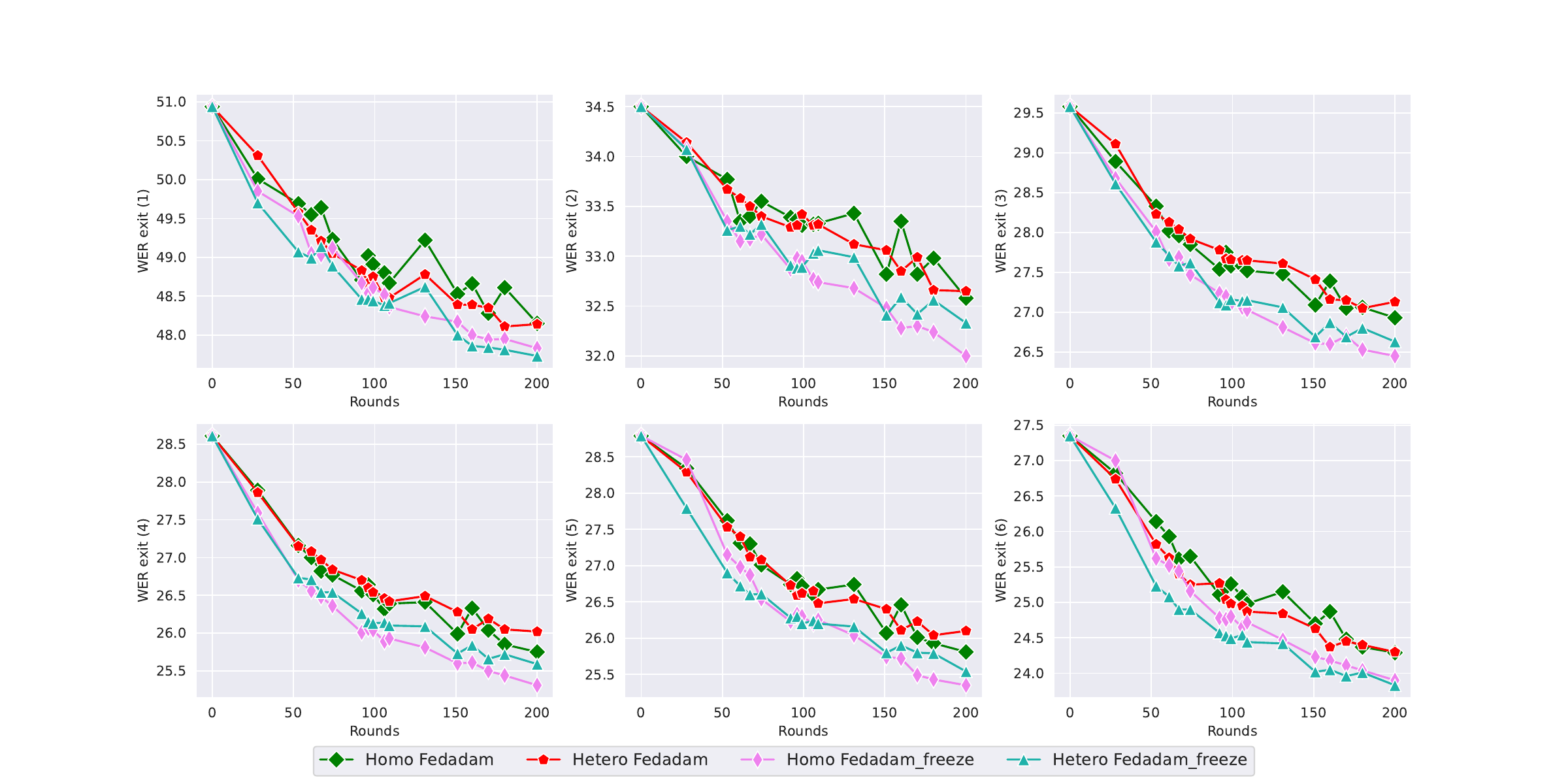}
    \caption{Comparison of the WER achieved with FedAdam on TED-LIUM-3 with (pink and pale colors) and without (red and green colors) freezing the convolutional front-end.}
  \label{fig:allresults}
\end{figure}

Fig.~\ref{fig:allresults} shows the beneficial impact of freezing the convolutional layers of the model: the pink and pale-blue lines are well below the other two, for all exits and both OFL and EFL. It is worth mentioning that freezing the convolutional front-end and training only the encoder-decoder part is a common practice in ASR when fine-tuning pre-trained models \cite{liu2021autofreeze,lee2019would}. 

Finally, Fig. \ref{fig:vox} supports our proposed approach  confirming the convergence trends observed so far also on VoxPopuli, for both EFL and OFL.

\begin{figure}[ht!]
  \centering
      \includegraphics[width=\textwidth]{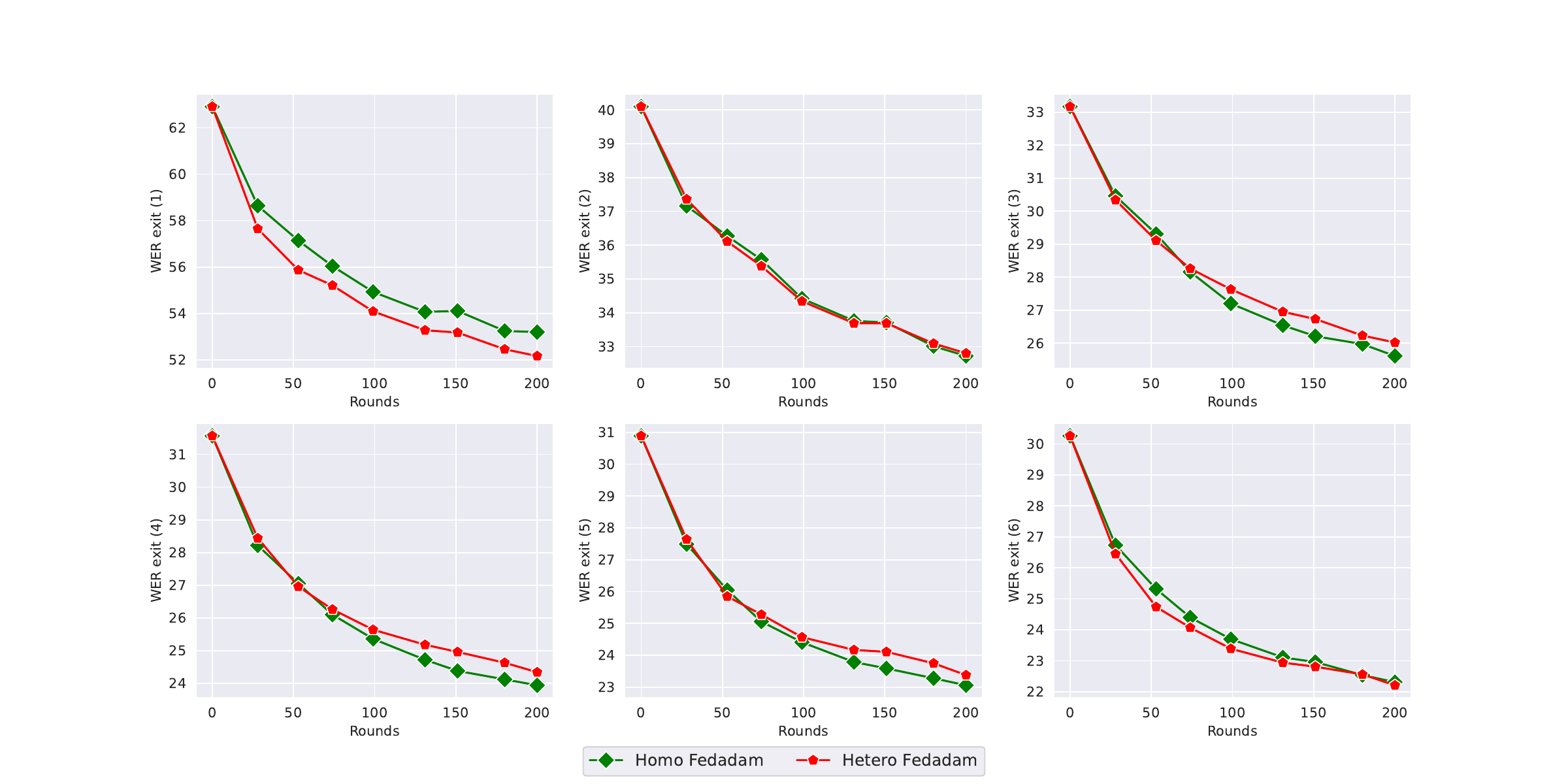}
    \caption{Comparison of the WER achieved on VoxPopuli with freezing the convolutional front-end for ELF and OFL.}
  \label{fig:vox}
\end{figure}

\subsection{Non-uniform distribution of heterogeneous clients}
\label{sub:expenonuniform}
The experiments reported above consider a rather ideal case where the different sub-nets $\w(\tau,l_m)$ are uniformly distributed across the clients. This implies that the central server receives enough updates for the upper exits. However, in  practical scenarios, we may expect that a large number of participating devices offer limited resources~\cite{ye2023heterogeneous}. This leads to an unbalanced device heterogeneity \cite{li2020federated}. To mimic this more realistic scenario, we simulate non-uniform distributions of the devices where the clients fitting the first exists are more frequent. This non-uniform distribution of clients can significantly impact the FL process as it affects the weights $\xi_m$ in Eq.~\ref{eq:m-dependent-lr} via the ratio $|\mathcal{C}_m|/|\mathcal{C}|$. As seen in Sec.~\ref{sec:proposed}, low probabilities for the upper exits feed the central server with few (or even none) observations for adapting the upper attention blocks of the encoder with the related linear decoders. In particular, we consider 2 scenarios using the probability distributions depicted in Fig. \ref{fig:dist} (from 1 to $M$): a {\it regular} distribution where 80\% of the clients cover the first 3 exits and an {\it extreme} distribution where the 80\% of the clients feature only the first exit.
\COMMENT{\begin{enumerate}
    \item regular: $[0.4, 0.2, 0.2, 0.1, 0.05, 0.05]$;
    \item extreme: $[0.8, 0.1, 0.025, 0.025, 0.025, 0.025]$
\end{enumerate}
}
\begin{figure}[ht!]
  \centering
    \includegraphics[width=0.7\textwidth]{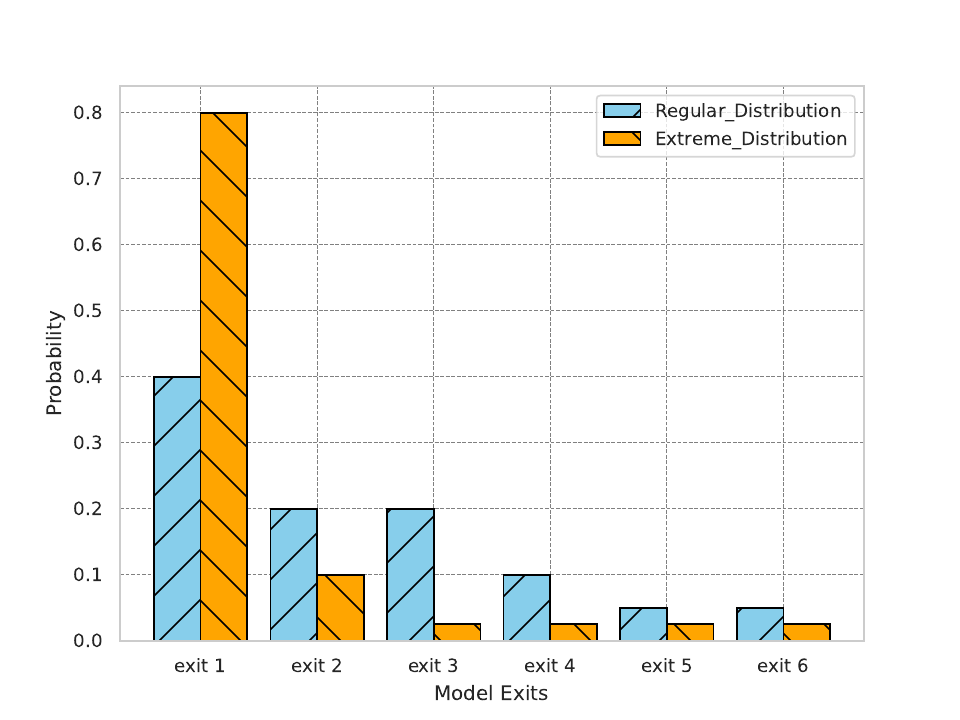}
    \caption{The regular and extreme non-uniform clients distributions.}
  \label{fig:dist}
\end{figure}
\begin{figure}[ht!]
  \centering
    \includegraphics[width=\textwidth]{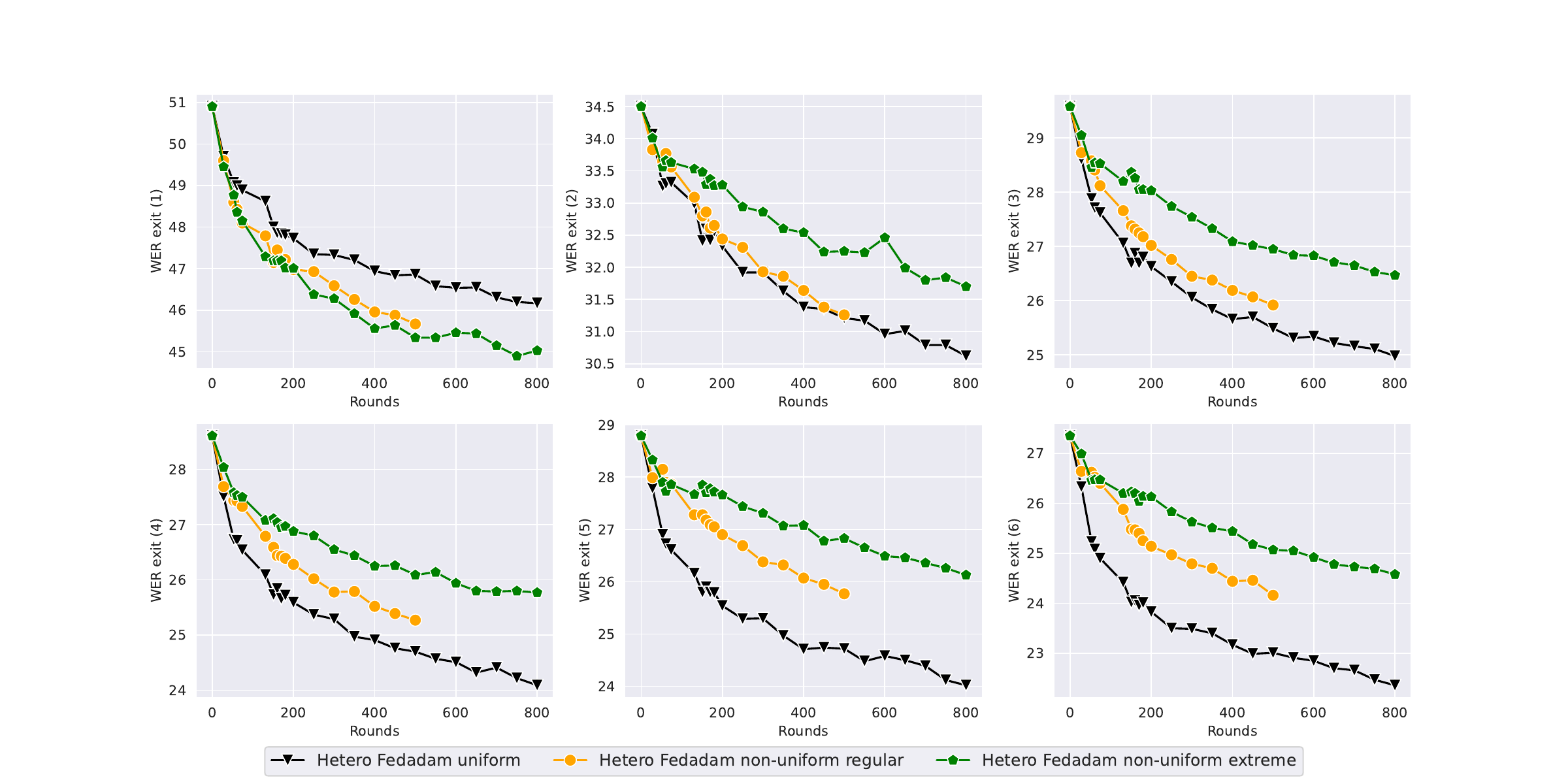}
    \caption{WER achieved with FedAdam in case of uniform and non-uniform heterogeneous clients distributions on TED-LIUM-3.}
  \label{fig:non-uniform}
\end{figure}
Figure~\ref{fig:non-uniform} reports the WER results across different FL rounds for the uniform (black) and the 2 non-uniform distributions (yellow and green).  
Note that, in the non-uniform scenario, the first exit converges faster than in the uniform case as $40\% $ of clients ($80\%$ in the extreme case) train this exit only, which is also trained in all other clients. Therefore, the training is heavily focused on improving that part of the network. Conversely, the model convergence in upper exits is definitely slower. Nevertheless, it is still possible to train those upper exists and improve the recognition performance. The figure confirms the fact that the non-uniform heterogeneous scenario leads to a sub-net dependent learning rate as in Eq.~\ref{eq:m-dependent-lr}.

These latter results may seem rather counter-intuitive: if the lower encoders are trained without accounting for the upper ones, we would expect that the performance of the upper exits deteriorates. Instead, it appears that 
improving the lower attention blocks is beneficial (or not detrimental) for the upper layers too, as long as the model is sufficiently trained. 
We experimentally verified this by performing central training on Librispeech and fine-tuning using only the exits lower than 2 ($m<2$) in the compound loss of Eq.~\ref{eq:EEloss}. The results are reported in Table~\ref{tab:centralnonuniform} for all of the 6 exits. We consider the following training conditions: {\em a)} training from scratch, {\em b)} fine-tuning from a model pre-trained on 960 hour of Librispeech and {\em c)} fine-tuning from a model pre-trained on 100 hours of Librispeech. For the latter 2 cases, the table reports the initial performance and the performance reached after fine-tuning for 100 epochs (the relative improvements are shown in parenthesis).
As expected training from scratch improves the performance of only the first two exits. Conversely, if the model is reasonably pre-trained, fine-tuning only the lower exits does not compromise the performance of the upper ones. 

\begin{table}[ht]\centering
\caption{Performance obtained on Librispeech test-clean data set with central training, starting from  different initial checkpoints (also from scratch) and fine-tuning only the first two exits of the EE model with SGD.}
\label{tab:centralnonuniform}
\scriptsize
\centering
\begin{tabular}{lc ccc cc}\toprule
& \textbf{\makecell{From \\ Scratch}} &\multicolumn{2}{c}{\textbf{\makecell{Librispeech\\ 960h}}} &\multicolumn{2}{c}{\textbf{\makecell{Librispeech \\ 100 h}}} \\\cmidrule{2-6}
\textbf{Exit} & fine-tuned & \makecell{Initial \\ Perform.} & fine-tuned & \makecell{Initial \\ Perform.} & fine-tuned \\ \cmidrule(l{2pt}r{2pt}){2-2}\cmidrule(l{2pt}r{2pt}){3-4} \cmidrule(l{2pt}r{2pt}){5-6}
\textbf{1} & 35.96 & 24.10  & 23.81 (1.2\%) & 42.28  & 41.15 (2.7\%) \\\cmidrule{1-6}
\textbf{2} & 20.40 & 11.78  & 11.67 (0.9\%) & 22.58  & 21.76 (3.6\%)  \\\hline\hline
\textbf{3} & 142.72 & 6.91  & 6.97 (-0.8\%) & 17.74  & 17.49 (1.4\%)  \\\cmidrule{1-6}
\textbf{4} & 102.44 & 6.28  & 6.15 (2\%)    & 15.58  & 15.26 (2\%)   \\\cmidrule{1-6}
\textbf{5} & 100.49 & 7.30  & 6.89 (5.6\%)  & 15.17  & 15.10 (0.4\%)  \\\cmidrule{1-6}
\textbf{6} & 100.35 & 5.34  & 5.24 (1.8\%)  & 14.73  & 14.61 (0.8\%)  \\\midrule

\end{tabular}
\end{table}

\section{Conclusions}
In this investigation, we present a comprehensive examination of federated learning within the context of heterogeneous devices, utilizing early-exit ASR architectures. Our study mathematically shows that the use of early-exit solutions with intermediates output layers and corresponding losses, allows federating heterogeneous models in a straightforward manner, obviating the necessity for specifically tailored agglomeration strategies. Through experimental evaluations conducted on the widely recognized TED-LIUM-3 and VoxPopuli benchmark datasets, we substantiate our claims showing the efficacy of early-exit architectures. Notably, we show experimentally that even in very unfavorable conditions, where the large majority of the clients updates only the lower part of the network, the convergence of the model is still achieved, although at a slower rate. Finally, through ablation studies we show that freezing the convolutional front-end of the pre-trained model in combination with FedAdam significantly enhances convergence in comparison to FedAvg.



\bibliographystyle{elsarticle-num}

\bibliography{bibsample.bib}

\end{document}